\documentclass[11pt,a4paper]{article}
\usepackage{authblk} 
\usepackage[hyperref]{emnlp2020}
\usepackage{times}
\usepackage{latexsym}

\usepackage{todonotes}
\usepackage{enumitem}
\usepackage{covington}
\usepackage{subcaption}

\usepackage{balance}

\usepackage{microtype}

\aclfinalcopy 
\def\aclpaperid{127} 

\newcommand{\stave}{\emph{Stave}}
\newcommand{\forte}{\emph{Forte}}

\title{
A Data-Centric Framework for Composable NLP Workflows
}

\renewcommand\Affilfont{\small}
\makeatletter
\renewcommand\AB@affilsepx{~~~~~~\protect\Affilfont}
\makeatother

\author[1,2,*]{Zhengzhong Liu}
\author[2]{Guanxiong Ding}
\author[2]{Avinash Bukkittu}
\author[2]{Mansi Gupta}
\author[2]{Pengzhi Gao}
\author[2]{Atif Ahmed}
\author[1]{Shikun Zhang}
\author[1,2]{\authorcr Xin Gao}
\author[2]{Swapnil Singhavi}
\author[1]{Linwei Li}
\author[1]{Wei Wei}
\author[1]{Zecong Hu}
\author[1]{Haoran Shi}
\author[3]{Haoying Zhang}
\author[3]{Xiaodan Liang}
\author[1]{\authorcr Teruko Mitamura}
\author[1,2]{Eric P. Xing}
\author[1,4]{Zhiting Hu}
\affil[1]{Carnegie Mellon University}
\affil[2]{Petuum Inc.}
\affil[3]{Sun Yat-sen University}
\affil[4]{UC San Diego\authorcr}
\affil[*]{\texttt{hectorzliu@gmail.com}}

\date{}

\begin{document}
\maketitle
\begin{abstract}
Empirical natural language processing (NLP) systems in application domains (e.g., healthcare, finance, education) involve interoperation among multiple components, ranging from data ingestion, human annotation, to text retrieval, analysis, generation, and visualization. We establish a unified open-source framework to support fast development of such sophisticated NLP workflows in a composable manner. The framework introduces a uniform data representation to encode heterogeneous results by a wide range of NLP tasks. It offers a large repository of processors for NLP tasks, visualization, and annotation, which can be easily assembled with full interoperability under the unified representation. The highly extensible framework allows plugging in custom processors from external off-the-shelf NLP and deep learning libraries.
The whole framework is delivered through two modularized yet integrable open-source projects, namely \forte{}\footnote{https://github.com/asyml/forte} (for workflow infrastructure and NLP function processors) and \stave{}\footnote{https://github.com/asyml/stave} (for user interaction, visualization, and annotation). 

\end{abstract}



\begin{figure*}[ht!] 
    \centering
    \includegraphics[width=\linewidth]{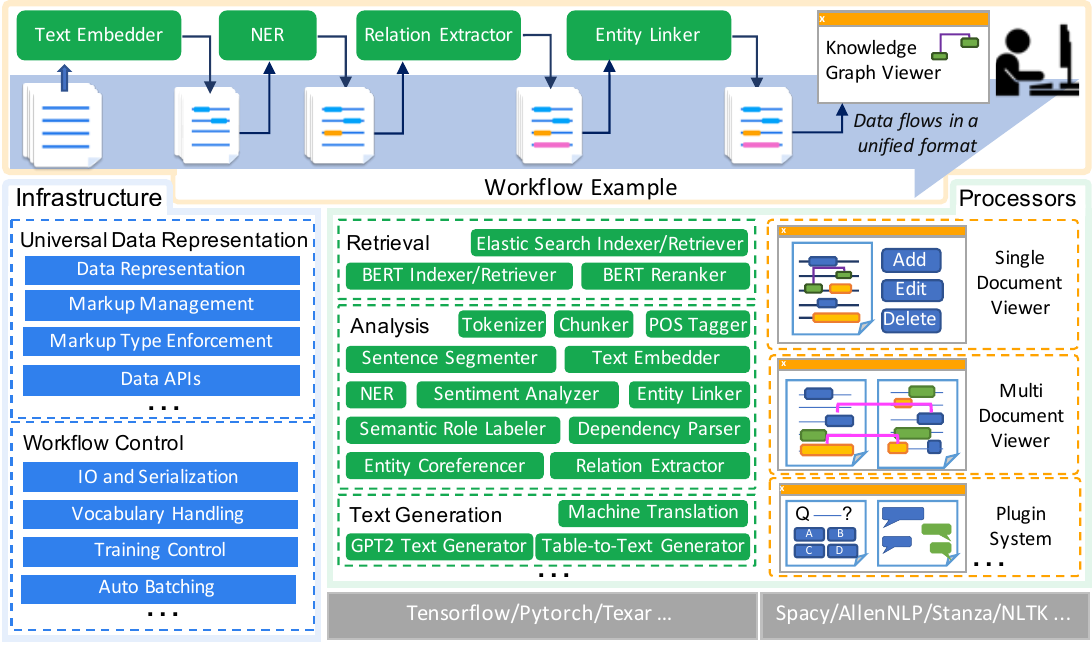}
    \vspace{-20pt}
    \caption{
    Stack of the data-centric framework for NLP workflows, including workflow \emph{infrastructure}, and \emph{processors} for NLP tasks and interactions (e.g., visualization, annotation). Different processors are composed together with the infrastructure APIs to form an arbitrary complex workflow. The example workflow transforms an unstructured text corpus into a knowledge graph through a series of NLP functions.
    }
    \label{fig:main_architecture}
    \vspace{-10pt}
\end{figure*}

\section{Introduction}
Natural language processing (NLP) techniques are playing an increasingly central role in industrial applications. A real-world NLP system involves a wide range of NLP tasks that interoperate with each other and interact with users to accomplish complex workflows. For example, in an assistive medical system for diagnosis (Figure~\ref{fig:medical_pipeline}), diverse text \emph{analysis} tasks (e.g., named entity recognition, relation extraction, entity coreference) are performed to extract key information (e.g., symptoms, treatment history) from clinical notes and link to knowledge bases; a medical practitioner could select any extracted entity to \emph{retrieve} similar past cases for reference; text \emph{generation} techniques are used to produce summaries from diverse sources.



To develop domain-specific NLP systems fast, it is highly desirable to have a unified open-source framework that supports:
(1) seamless integration and interoperation across NLP functions ranging from text analysis to retrieval to generation; (2) rich user interaction for data visualization and annotation; (3) extensible plug-ins for customized components; and (4) highly reusable components.

A wealth of NLP toolkits exist (\S\ref{sec:related}), such as spaCy~\citep{Honnibal2017}, DKPro~\citep{EckartdeCastilho2014},
CoreNLP~\citep{Manning2014},
for pipelining multiple NLP functions; BRAT~\citep{Stenetorp2012} and YEDDA~\citep{Yang2018} for annotating certain types of data. None of them have addressed all the desiderata uniformly. Combining them for a complete workflow requires non-trivial effort and expertise (e.g., ad-hoc gluing code), posing challenges for maintenance and upgrading.

We introduce a new unified framework to support complex NLP workflows that involve text data ingestion, analysis, retrieval, generation, visualization, and annotation. The framework provides an infrastructure to simply plug in arbitrary NLP functions and offers pre-built and reusable components to build desired workflows. Importantly, the framework is designed to be extensible, allowing users to write custom components (e.g., specialized annotation interfaces) or wrap other existing libraries~\citep[e.g.,][]{hu2019texar,wolf2019huggingface} easily.



The framework's design is founded on a data-centric perspective. We design a universal text data representation that can encode diverse input/output formats of various NLP tasks uniformly. Each component (``\emph{processor}'') in the workflow fetches relevant parts of data as inputs, and passes its results to subsequent processors by adding the results to the data flow (Figure~\ref{fig:main_architecture}). In this way, different processors are properly decoupled, and each is implemented with a uniform interface without the need of accommodating other processors. Visualization and annotation are also abstracted as standalone components based on the data representation.

We demonstrate two case studies on using the framework to build a sophisticated assistive medical workflow and a neural-symbolic hybrid chatbot.



\begin{figure*}
    \centering
    \includegraphics[width=0.9\textwidth]{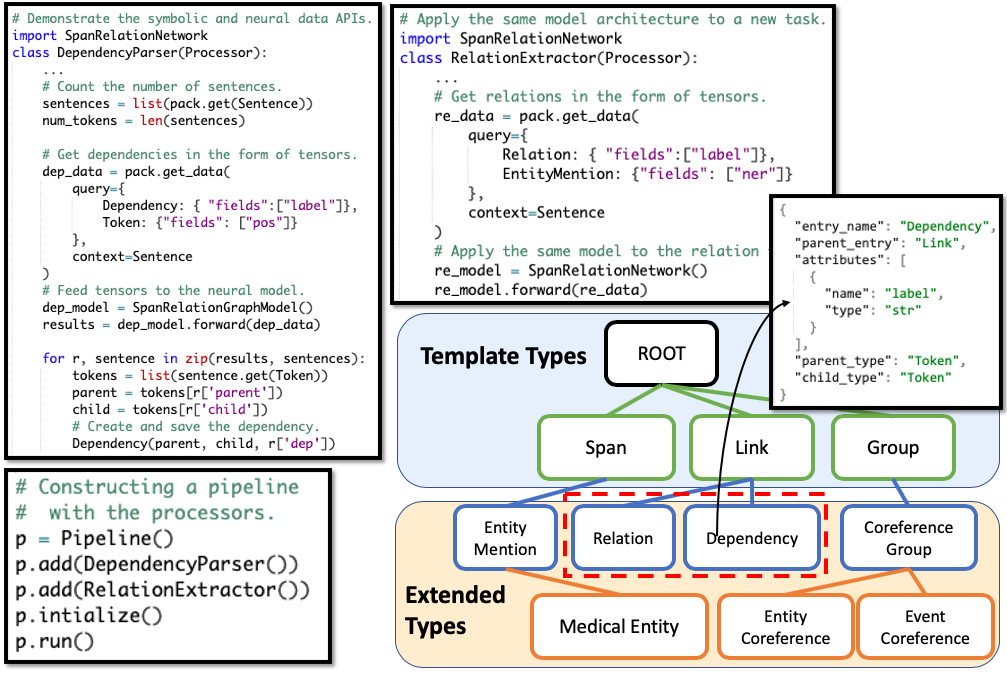}
    \vspace{-10pt}
    \caption{\textbf{Top Left}: A dependency parser processor that calls a neural model and save the results; \textbf{Top Right}: A relation extractor can use the same model architecture. \textbf{Bottom Left}: A pipeline can be constructed by simply adding processors. \textbf{Bottom Right}: Example data types offered by the framework or customized by users. \texttt{Relation} and \texttt{Dependency} both extends \texttt{Link}. Definition of \texttt{dependency} is done through a simple JSON configuration.}
    \label{fig:ontology}
    \vspace{-10pt}
\end{figure*}

\section{Data-Centric NLP Framework}\label{sec:workflow}

Figure~\ref{fig:main_architecture} shows the stack of the framework, consisting of several major parts: {\bf (1)} We first introduce the underlying infrastructure (\S\ref{sec:infra}), in particular, a universal representation scheme for heterogeneous NLP data. The highly-organized unified representation plays a key role in supporting composable NLP workflows, which differentiates our framework from prominent toolkits such as CoreNLP~\citep{Manning2014}, spaCy~\citep{Honnibal2017}, and AllenNLP~\citep{Gardner2019}. We then introduce a range of functionalities that enable the convenient use of the symbolic data/features in neural modeling, which are not available in traditional NLP workflow toolkits such as DKPro~\citep{EckartdeCastilho2014}.
{\bf (2)} \S\ref{sec:language_processing} describes how processors for various NLP tasks can be developed with a uniform interface, and can be simply plugged into a complex workflow. {\bf (3)} finally, the human interaction part offers rich composable processors for visualization, annotation, and other forms of interactions.


\subsection{Infrastructure}\label{sec:infra}

\subsubsection{Universal Data Representation}\label{sec:data}
\begin{figure*}[ht!] 
    \centering
    \includegraphics[width=0.9\textwidth]{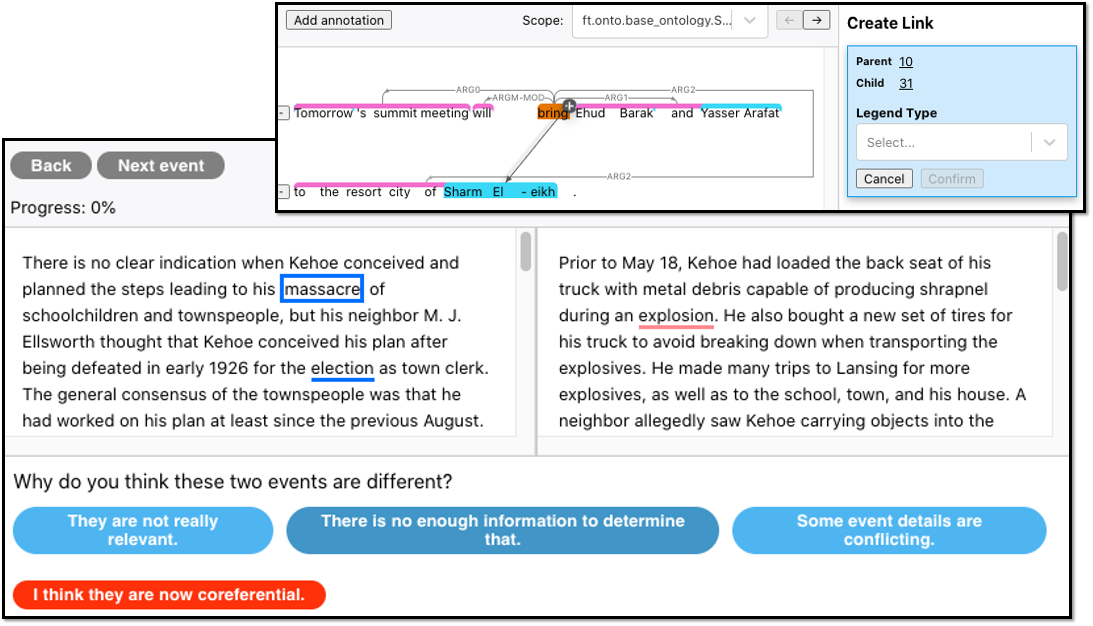}
    \vspace{-10pt}
    \caption{\textbf{Top}: Screenshot of the single doc interface shows predicates, entity mentions and semantic role links of one sentence. A new link is being created from ``bring" to ``Sharm Ei-eikh". \textbf{Bottom}: Screenshot of the two document interface for annotating event coreference. The system is suggesting a potential coreference pair. The interfaces are rendered based on the data types. Users can customize the interface to use different UI components.}
    \label{fig:stave_screen_shot}
    \vspace{-10pt}
\end{figure*}

NLP data primarily consists of two parts: the raw \emph{text source} and the structured \emph{markups} on top of it (see Figure~\ref{fig:stave_screen_shot} for an example).
The markups represent the information overlaid on the text, such as part-of-speech tags, named entity mentions, dependency links, and so forth. NLP tasks are to produce desired text or markups as output, based on vastly different input information and structures,

To enable full interoperation among distinct tasks, we summarize the underlying commonalities between the myriad formats across different NLP tasks, and develop a universal data representation encapsulating information uniformly. 
The representation scheme defines a small number of \emph{template} data types with high-level abstraction, which can be further extended to encode domain-specific data.

\paragraph{Template data types:}
We generalize the previous UIMA representation scheme~\citep{Gotz2004} to cover the majority of common NLP markups. This results in three template data types, each of which contains a couple of attributes.
%
\begin{itemize}[noitemsep,leftmargin=10pt]
    \item \texttt{Span} contains two integer attributes, \textit{begin} and \textit{end}, to denote the offsets of a piece of text. This type can mark tokens, entity mentions, and etc.
    \item \texttt{Link} defines a pair of (\textit{parent}, \textit{child}) which are pointers to other markups, to mark dependency arcs, semantic roles, entity relations, etc.
    \item \texttt{Group} has a \textit{members} attribute, which is a collection of markups. This type can mark coreference clusters, topical clusters, etc.
\end{itemize}


\paragraph{Extended data types:}
In order to encode more specific information, each of the template data types can be extended by adding new attributes. For example, the framework offers over 20 extended types for commonly used NLP concepts, such as \texttt{Token} and  \texttt{EntityMention}. Moreover, users can easily add \emph{custom} data types through simple JSON definitions (Figure~\ref{fig:ontology}) to fulfill specific needs, such as \texttt{MedicalEntity} that extends \texttt{EntityMention} with more attributes like patient IDs. 
Once a new data type is defined, rich data operations (e.g., structured access) as below are automatically enabled for the new type. 

\paragraph{Flexible Data Sources:}
Modern NLP systems face challenges imposed by the volume, veracity and velocity of data. To cope with these, the system is designed with customizable and flexible data sources that embrace technologies such as Indexing (e.g. Elastic Search~\cite{Elastic.co}), Databases (e.g. Sqlite), Vector Storage (e.g. Faiss~\cite{JDH17}). Users are free to implement flexible ``Reader" interface to ingest any source of data. 


%

\subsubsection{Facilitation for Neural Modeling}\label{sec:neural_symbolic}
%
The framework provides extensive functionalities for effortless integration of the above symbolic data representation with tensor-based neural modeling.

\textbf{Neural representations.} All data types are associated with an optional \emph{embedding} attribute to store continuous neural representations. Hence, users can easily access and manipulate the embeddings of arbitrary markups (e.g., entity, relation) 
extracted from neural models like word2vec~\cite{Mikolov2013EfficientEO} and BERT~\cite{devlin2019}.
The system also supports fast embedding indexing and lookup with embedding storage systems such as Faiss~\cite{JDH17}.

\textbf{Rich data operations: auto-batching, structured access, etc. }
Unified data representation enables a rich array of operations to support different data usage, allowing users to access any information in a structured manner. Figure~\ref{fig:ontology} (top left) shows 
API calls that get all dependency links in a sentence.
Utilities such as \emph{auto-batching} and \emph{auto-padding} help aggregates relevant information (e.g., event embeddings) from individual data instances into tensors, which are particularly useful for neural modeling on GPUs. 



\textbf{Neural-symbolic hybrid modeling.} 
Unified data representations and rich data operations make it convenient to support hybrid modeling using both neural and symbolic features. Take retrieval for example, the framework offers retrieval processors (\S\ref{sec:language_processing}) that retrieve a coarse-grained candidate set with symbolic features (e.g., TF-IDF)  first, and then refine the results with more expensive embedding-based re-ranking~\cite{Nogueira2019PassageRW}. Likewise, fast embedding-based search is facilitated with the Faiss library~\cite{JDH17}.


\textbf{Shared modeling approaches. } 
The uniform input/output representation for NLP tasks makes it easy to share the modeling approaches across diverse tasks. For example, similar to \citep{Jiang2020}, all tasks involving the \texttt{Span} and \texttt{Link} data types as outputs (e.g., dependency parsing, relation extraction, coreference resolution) can potentially use the exact same neural network architecture for modeling. Further with the standardized APIs of our framework, users can spawn models for all such tasks using the same code with minimal edits. Top right of Figure~\ref{fig:ontology} shows an example where the same relation extractor is implemented with dependency parser for a new task, and the only difference lies in accessing different data features.


\subsection{Processors}\label{sec:language_processing}
Universal data representation enables a uniform interface to build processors for different NLP tasks. Most notably, \emph{interoperation} across processors supported by the system abstraction allows each processor to interacts with each other via the data flow.


Each processor takes uniformly represented data as inputs
and performs arbitrary actions on them. A processor can edit text source (e.g., language generation), add additional markups (e.g., entity detection), or produce side effects (e.g., writing data to disk). Top left of Figure~\ref{fig:ontology} shows the common structure of a processor, that fetches relevant information from the data pack with high-level APIs, performs operations such as neural model inference, writes results back to the data pack and pass them over to subsequent processors. Top right shows the simple API used for plugging processors into the workflow.





\textbf{A comprehensive repository of pre-built processors.} 
With the standardized concept-level APIs for NLP data management, users can easily develop any desired processors. One can wrap existing models from external libraries by conforming to the simple interfaces. Moreover, we offer a large set of pre-built processors for various NLP tasks, ranging from text retrieval, to analysis and generation. Figure~\ref{fig:main_architecture} lists a subset of processors.
\subsection{Visualization, Annotation, \& Interaction}
The interfaces for visualization and annotation are implemented as standalone components and designed for different data types. 

\textbf{Single document viewer. } We provide a single document interface (Figure~\ref{fig:stave_screen_shot}) that renders the template types. 
For example, Spans are shown by colored highlights, Links are shown as connectors between the spans. 
A user can create new spans, add links, create groups, or edit the attributes through intuitive interfaces. 


\begin{figure*}[ht!]
    \centering
    \includegraphics[width=0.9\linewidth]{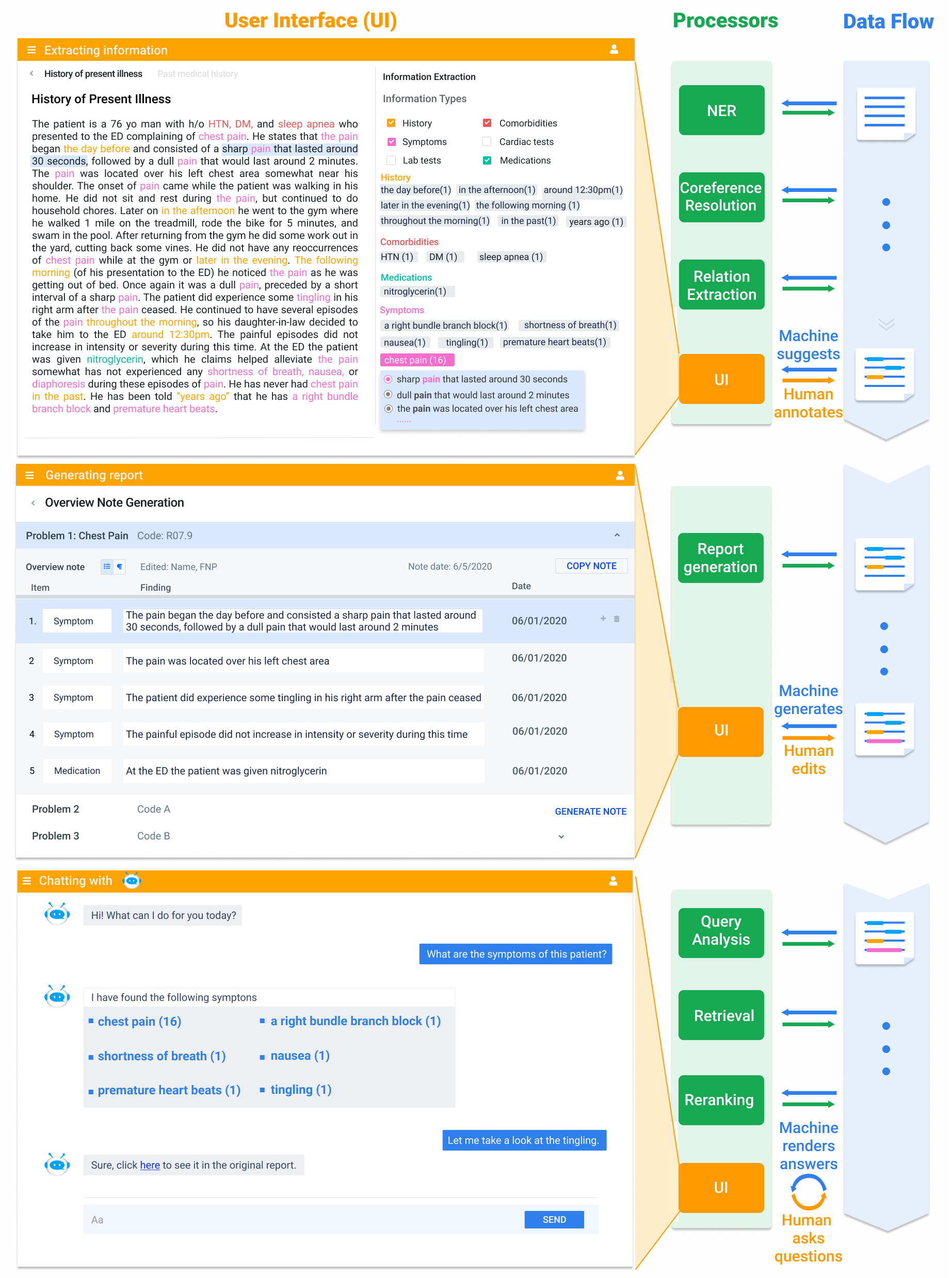}{}
    \vspace{-10pt}
    \caption{A system for diagnosis analysis and retrieval from clinical notes. The data-centric approach makes it easy to assemble a variety of components and UI elements. Example text was obtained from \newcite{UNCSchoolofMedicine}.}
    \label{fig:medical_pipeline}
    \vspace{-10pt}
\end{figure*}

\textbf{Multi document viewer. } \stave{} currently supports a two-document interface. Users can create links across documents. The bottom of Figure~\ref{fig:stave_screen_shot} shows an example of annotating event coreference. The system is suggesting an event coreference pair and asking for the annotator's decision.

\textbf{Customization with plugins. } 
While default interfaces support a wide range of tasks, users can create customized interfaces to meet more specific needs.
%
We build a system that can quickly incorporate independently-developed plugins, such as a plugin for answering multiple-choice questions, or a dialogue interface for chat-bots.
Some pre-built plugins are showcased in Figure~\ref{fig:medical_pipeline}.
Additionally, the layout can be customized to display specific UI elements, allowing for greater flexibility to use plugins together.

\textbf{Human-machine collaboration. }
Universal data representation across all modules not only enhances interoperation between components, but also allows machines and humans to collaborate in a workflow. Human-interactive components can be integrated at any part of the workflow for visualization/reviewing to produce a downstream system that combines the advantages of humans and machines.
Machine-assisted annotation can be undertaken straightforwardly: the annotation system simply ingests the data produced by a back-end processor (Figure~\ref{fig:stave_screen_shot}).

\section{Case Studies}
\subsection{A Clinical Information Workflow}\label{sec:case_study}
We demonstrate an information system for clinical diagnosis analysis, retrieval, and user interaction.
Figure~\ref{fig:medical_pipeline} shows an overview of the system.
To build the workflow, we first define domain-specific data types, such as \texttt{Clinical Entity Mention}, via JSON config files as shown in Figure~\ref{fig:ontology}. We then develop processors for text processing: (1) we create an LSTM-based clinical NER processor~\cite{Boag2015}, a Span-Relation model-based relation extraction processor~\cite{He2018}, and a coreference processor with the end-to-end model~\cite{Lee2017} to extract key information; (2) we build a report generation processor following \newcite{Li2019} with extracted mentions and relations; (3) we build a simple keyword based dialogue system for user to interact using natural languages.
The whole workflow is implemented with minimal engineering effort. For example, the workflow is assembled with just 20 lines of code; and the IE processors are implemented with around 50 lines of code by reusing libraries and models.



\subsection{A ChatBot Workflow}\label{sec:chatbot}
The case study considers the scenario where we have a corpus of movie reviews in English to answer complex queries (e.g., {``movies with a positive sentiment starring by a certain actor''}) by a German user.
The iterative workflow consists of a review retrieval processor based on the hybrid symbolic-neural feature modeling (\S\ref{sec:neural_symbolic}), an NER processor~\cite{Gardner2019} to find actors and movies from the retrieved reviews, a sentiment processor~\cite{Hutto2014VADERAP} for sentence polarity, and an English-German translation processor.

\section{Related Work}\label{sec:related}
The framework shares some characteristics with UIMA~\cite{Gotz2004} backed systems, such as DKPro~\cite{EckartdeCastilho2014}, ClearTK~\cite{bethard-etal:2014:LREC} and cTakes~\cite{ctakes}. 
There are NLP toolboxes like NLTK~\cite{Bird2016} and AllenNLP~\cite{Gardner2019}, GluonNLP~\cite{Guo2019}, NLP pipelines like Stanford CoreNLP~\cite{Manning2014}, SpaCy~\cite{Honnibal2017}, and Illinois Curator~\cite{Clarke2012}. 
As in \S\ref{sec:language_processing}, our system develops a convenient scaffold and provides a rich set of utilities to reconcile the benefits of symbolic data system, neural modeling, and human interaction, making it suitable for building complex workflows.


Compared to open-source text annotation toolkits, such as Protégé Knowtator~\cite{Ogren2006}, BRAT~\cite{Stenetorp2012}, Anafora~\cite{Chen2013}, GATE~\cite{Cunningham2013}, WebAnno~\cite{Castilho2016}, and YEDDA~\cite{Yang2018}, our system provides a more flexible experience with customizable plug-ins, extendable data types, and full-fledged NLP support. The tool Prodigy by spaCy is not 
open-source and supports only pre-defined annotation tasks like NER.


\section{Conclusions and Future Work}
We present a data-centric framework for building complex NLP workflows with heterogeneous modules.
We will continue to improve the framework 
on other advanced functionalities, such as multi-task learning, joint inference, data augmentation, and provide a broader arsenal of processors to help build better NLP solutions and other data science workflows. We also plan to further facilitate workflow development by providing more flexible and robust data management processors.



\balance
\bibliography{emnlp2020}
\bibliographystyle{acl_natbib}
\end{document}